\documentclass[11pt,a4paper]{article}
\usepackage[hyperref]{ranlp2023}
\usepackage{times}
\usepackage{latexsym}
\usepackage{graphicx}

\usepackage{amsmath}
\usepackage{amsfonts}
\usepackage{amssymb}
\usepackage{bbm}
\usepackage{multirow}

\usepackage{microtype}

\aclfinalcopy 

\def\q{\mathbf{q}}
\def\h{\mathbf{h}}
\def\e{\mathbf{e}}
\newcommand{\eref}[1]{Equation~(\ref{#1})}
\newcommand{\fref}[1]{Figure~\ref{#1}}

\title{Unsupervised Calibration through Prior Adaptation \\ for Text Classification using Large Language Models}

\author{
$^{\ast\star}$Lautaro Estienne, $^\ast$Luciana Ferrer, $^{\star\dagger}$Matías Vera, $^{\ddagger}$Pablo Piantanida \\
$^\ast$Instituto de Investigación en Ciencias de la Computación, CONICET-UBA, Argentina \\
$^\star$ Departamento de Electrónica, Facultad de Ingeniería, Universidad de Buenos Aires, Argentina \\
$^\dagger$ Centro de Simulación Computacional para Aplicaciones Tecnológicas, CONICET, Argentina \\
$^\ddagger$ Laboratoire des Signaux et Systèmes (L2S), CentraleSupelec CNRS Universite Paris-Saclay \\
\texttt{\{lestienne,mvera\}@fi.uba.ar} ~
\texttt{lferrer@dc.uba.ar} \\
\texttt{pablo.piantanida@centralesupelec.fr}
}

\newbool{showappendix}
\setbool{showappendix}{true}

\begin{document}
\maketitle
\begin{abstract}
A wide variety of natural language tasks are currently being addressed with large-scale language models (LLMs). These models are usually trained with a very large amount of unsupervised text data and adapted to perform a downstream natural language task using methods like fine-tuning, calibration or in-context learning. In this work, we propose an approach to adapt the prior class distribution to perform text classification tasks without the need for labelled samples and only few in-domain sample queries. 
The proposed approach treats the LLM as a black box, adding a stage where the model posteriors are calibrated to the task.  Results show that these methods outperform the un-adapted model for different number of training shots in the prompt and a previous approach were calibration is performed without using any adaptation data.
\end{abstract}

\section{Introduction}\label{sec:intro}
In the last years, Large Language Models (LLMs) like GPT-3~\citep{gpt3}, FLAN-T5~\citep{flant5}, InstructGPT~\citep{instructgpt} have proven to be useful for a large variety of complex natural language understanding tasks, showing outstanding performance on many benchmarks related to reading comprehension, summarization, information retrieval, and generative question-answering, among others~\citep{xsum,hellaswag,dsp,omar2023chatgpt}. LLMs are pre-trained on a large amount of unsupervised text data following a cost function that is usually self-supervised (autoregressive, denoising, etc.)~\citep{xlnet,flant5,bert}.

Notably, LLMs achieve competitive results in a zero-shot scenarios, i.e., without being adapted to the downstream task of interest~\citep{wei2021,flant5,gpt4}. Nevertheless, when data is available for adaptation, significant gains can be achieved over the zero-shot scenario. In these cases, the adaptation is done, for example, through (full or selective) fine-tuning~\citep{bert,flant5}, in-context learning~\citep{gpt3,wei2021}, or post-processing of the model's outputs~\citep{zhao2021calibrate,whenlm}, depending on the size of adaptation dataset, whether this data is labelled or not, and the amount of computational resources available. 

In this work, we propose an approach to adapt LLMs to text classification tasks using unlabelled in-domain data. That is, we assume we have examples of the type of text that needs to be classified, but we do not have the actual class of these examples. We propose a light-weight method inspired by the theory on calibration which, for the datasets we experimented with, required only a few dozen of in-domain samples to achieve optimal performance. We call this method \textbf{UCPA} (\textbf{U}nsupervised \textbf{C}alibration through \textbf{P}rior \textbf{A}daptation). We compare our proposed approach with a previously proposed approach which does not rely on any in-domain data \cite{zhao2021calibrate} and show that the additional information provides significant performance improvements. Further, we compare our method, theoretically and empirically,  with supervised calibration of the posteriors  using logistic regression. We show that our approach performs similarly to supervised calibration, without the need for labelled data. Finally, another version of the method is presented where we assume that, even though no labelled data is available, the class priors can be estimated from knowledge of the task.  We call this variant \textbf{SUCPA} (\textbf{S}emi-\textbf{U}nsupervised \textbf{C}alibration through \textbf{P}rior \textbf{A}daptation) since some information about the task is needed to estimate the priors.


\section{Related Work}

\paragraph{Large Language Models (LLMs)} LLMs are language models with a large number of parameters, in the order of billions, trained with a massive amount of text to minimize a cost function that can vary from model to model. Models like GPT-2~\citep{gpt2}, GPT-3~\citep{gpt3} or LLaMA~\citep{llama} are decoder-only transformer-based~\citep{transformer} architectures trained with an autoregressive loss. In contrast, models like BERT~\citep{bert} or T5~\citep{t5} are trained to denoise the input to obtain the output. 

\paragraph{Fine-tuning} Pre-trained LLMs can be adapted to a specific task of interest using finetuning techniques like Parameter Efficient Finetuning (PEFT)~\citep{peft}, soft-prompt~\citep{prompttuning} and Reinforcement Learning from Human Feedback (RLHF)~\cite{instructgpt}. 
Despite the attempt of some methods to reduce the number of trainable parameters without sacrifice performance, finetuning is generally an expensive way of adapting a LLM to a certain task since it requires significant amounts of in-domain data, as well as computational resources to load and train the LLM. 

\paragraph{In-context Learning} Given the large computational and data requirements of the fine-tuning approach, alternative approaches to adapt LLMs to a certain task of interest have been proposed. In-context learning refers to the practice of pre-pending instructions about the task of interest before the text to be classified, summarized or continued in some other way. Besides these instructions, examples (usually called ``shots'') on how to perform the task can be added to the prompt. The GPT-3 paper~\citep{gpt3} showed that close to the state-of-the-art performance in many tasks could be obtained by providing instructions in the prompt without any further adaptation to the task. This work led to the study of good prompt design practices~\citep{zhao2021calibrate}.

\paragraph{Calibration} There is a large body of literature regarding calibration of classifiers' outputs~\citep{filho2021}, with some recent applications to  Natural Language Processing Tasks~\citep{jagannatha2020,braverman2019}. Recently, a work from \citet{zhao2021calibrate} used the concept of ``content-free'' input to perform an ad-hoc unsupervised form of calibration for different tasks carried out by a LLM. Our work can be seen as a generalization and formalization of this work, where we derive the approach as unsupervised calibration with an affine expression where the parameters are obtained through minimization of the cross-entropy.

\section{UCPA: Unsupervised Calibration through Posterior Adaptation}\label{sec:ucpa}

\begin{figure*}[t!]
    \centering
   \includegraphics[width=.9\textwidth]{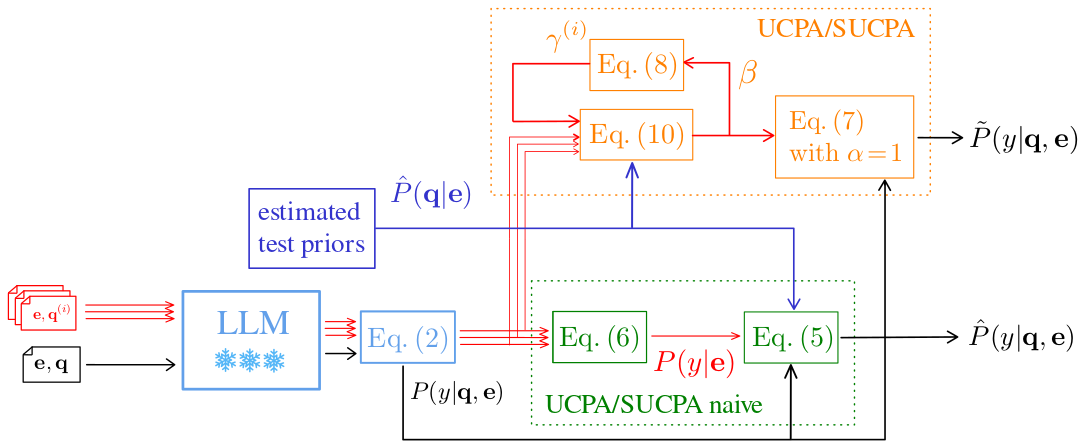}
    \caption{Schematic of the proposed approach. The test sample $(\e,\q)$ is processed by the LLM and plugged to \eref{eq:post_from_llm} to produce the posterior $P(y \mid \q, \e)$. In addition, a set of in-domain queries $\{\q^{(1)},\ldots,\q^{(N)}\}$ is used to reestimate the priors in a ``naive'' (bottom) or iterative (up) way. Lastly, estimated test priors $P(\q\mid \e)$ are used to produce adapted posteriors $\tilde P(y\mid \q,\e)$ or $\hat P(y\mid \q,\e)$. For the UCPA approach, $P(\q\mid \e)$ is assumed uniform, whereas for SUCPA, specific knowledge of the task is used to estimate that prior.}
    \label{fig:architecture}
\end{figure*}

An LLM produces posterior probabilities $P(n | \h, \q, \e)$ for the next token, $n$, given the history of previously generated words or tokens, $\h$, and the query, $\q$, and preface, $\e$, which together form the prompt $(\e, \q)$. 
The bolded variables indicate sequences of one or more tokens, while $n$ is a single token. In our case, the preface contains instructions on the classification task to be solved~\citep{flant5} and, optionally, a set of training examples for in-context learning~\citep{gpt3}.

When using a LLM to do classification, we need to use the model to obtain $P(y | \q,\e)$ from a prompt $(\e, \q)$, where $y \in \mathcal{Y}$ and $\mathcal{Y}=\{y_1, \ldots, y_K\}$ is the set of possible classes. To do this, we first define an \textit{ad-hoc} label name $w_k$ for every label $y_k \in \mathcal{Y}$. Then, we prompt the LLM, which we will call $\theta$, with the word sequence given by $\e$ followed by $\q$ to get a score 
\begin{equation}
s_k=P_{\theta}(w_{k}|\q,\e)
\end{equation}
for the class $y_k$. Finally, the probability distribution $P(y|\q,\e)$ is computed by normalizing this score to get a probability distribution over the classes:
\begin{equation}
P(y = y_k|\q,\e)=\frac{s_k}{\sum_{k'} s_{k'}}\label{eq:post_from_llm}
\end{equation}
%

Note that there may be cases in which the label name is represented with more than one token (see Table \ref{tab:datasets} for a complete list of datasets used and the label names of each one). In those cases the probability $P_\theta(w_{k}|\q,\e)$ can be computed as
\begin{eqnarray}\label{eq:chain_lm}
P_\theta(w_k|\q,\e) = \prod_{m=0}^{M_k-1} P_\theta(w_k^{m+1} \mid \q, \e, w_{k}^{1:m}) 
\end{eqnarray}
where $w_k^{1:m} = [w_k^{1}, \ldots, w_k^{m}]$ and $w_k^{1:0}$ is an empty string. 
The posteriors in the right-hand side are obtained directly from the LLM. 

Using the definition of conditional probability, the posterior $P(y | \q, \e)$ can be written as:
\begin{align}\label{eq:post_decomposition}
P(y \mid \q, \e)  =  P(y  \mid \e) \ \frac{P(\q \mid y, \e)}{P(\q \mid \e)} 
\end{align}
The factor $P(\q \mid y, \e)/P(\q \mid \e)$ is the ratio between the likelihood of the query given the preface, and the class name, and the likelihood given only the preface. This likelihood ratio (LR) reflects the increase in likelihood of the query obtained by adding the class name to the response. 

The quantity $P(y | \e)$ can be understood as a prior probability in the sense that it is not conditioned on the query: it is the probability of the class given only the preface. This prior depends strongly on the task of interest. While the LLM might have a tendency to predict a certain class given the preface, this may not be the most likely class for the task of interest. As we mentioned in Section \ref{sec:intro}, adapting the model to the application of interest is key for obtaining relevant responses from the LLM. 
While the preface $\e$ is, in fact, a way to adapt the LLMs outputs to the task of interest, it may not be sufficient to fully adapt the posteriors since not all the information about the task can be represented in a short text explanation.

In this work we propose to improve the posteriors computed from the LLM's scores by explicitly adjusting the priors to the task of interest. This is done by assuming the following expression for the in-domain posterior:
\begin{equation}\label{eq:new_post}
\hat P(y\mid  \q, \e)  = \delta \ P(y\mid\q,\e) \frac{\hat P(y\mid\e)}{P(y \mid \e)}
\end{equation}
which can be interpreted as taking away the effect of the mismatched prior $P(y | \e)$ from the LLM-derived posterior $P(y | \q,\e)$, replacing it with the in-domain prior $\hat P(y | \e)$, and then rescaling by $\delta$ to make sure the resulting distribution adds up to one. 
To obtain $\hat P(y | \q, \e)$ we need to compute the posterior, which is obtained directly from the LLM using \eref{eq:post_from_llm}, and the two priors. 

The prior $\hat P(y|\e)$ is the prior we expect for our task of interest. 
We may or may not know this distribution. In this work we compare results under two assumptions: 1) that we do not know anything about the prior distribution in which case we simply assume a uniform distribution $\hat P(y_k|\e) = 1/K$ for all $k$, and 2) that, even though we do not have labelled data, we do have a good estimate of the frequencies of the classes we expect to see in practice. In our experiments, for this second scenario we compute $\hat P(y_k|\e) = N_k/N$, where $N_k$ is the number of training samples of class $k$. In practice, though, these priors could be estimated from knowledge of the task rather than from in-domain labelled data. Arguably, this second scenario is no longer unsupervised, so we will call this method Semi-Unsupervised Calibration through Prior Adaptation (SUCPA), while the method that uses uniform priors will be called Unsupervised Calibration through Prior Adaptation (UCPA). 

The prior $P(y\mid\e)$ is the prior for $y$ that is implicit in our LLM. It is the distribution of classes that the model would output for this task, across all possible relevant queries we may provide. Hence, we estimate this prior by simply running the LLM on (unlabelled) training data $Q^\mathrm{train}=\{\q^{(1)},\ldots,\q^{(N)}\}$ and averaging the resulting posteriors:
\begin{equation}\label{eq:prior_approx}
    P(y|\e) \approx \frac{1}{N}\sum_{i=1}^N P(y|\q^{(i)},\e)
\end{equation}
The method above is a heuristic that relies on an assumption (\eref{eq:new_post}) that may or may not hold, as well as on the approximation of the prior above. When using this expression to obtain the prior we call the method ``UCPA/SUCPA-naive''. As we will see, this heuristic works quite well in our experiments. Further, as we explain in Section \ref{sec:calibration}, we can also obtain the prior in a more principled way using an expression derived from linear logistic regression.



\subsection{Content-free Prompts}\label{sec:content_free}

The approach proposed in \citet{zhao2021calibrate} can be seen as a special case of the UCPA-naive method. In that work, calibrated posteriors are computed using \eref{eq:new_post}, where the training set $Q^\mathrm{train}$ is composed of one or more ``content-free'' inputs, in the sense that they do not contain any relevant meaning, and they are created manually by the user. For example, the authors experiment with using ``[MASK]'', ``N/A'', and the empty string.

\section{Supervised Affine Calibration and Semi-UCPA (SUCPA)}
\label{sec:calibration}
A standard way to adapt the posterior probabilities from a classifier to a certain domain of interest is to calibrate them using in-domain labelled data. Calibration refers the process of transforming the scores of a system to optimize the quality of the scores as posteriors. This is usually done by choosing a certain parameterized form for the transform and training those parameters to minimize a proper scoring rule like the cross-entropy~\cite{filho2021}. One instance of this approach is linear logistic regression.

Linear logistic regression assumes that the logarithm of the calibrated posteriors are given by:
\begin{equation}\label{eq:cal_post}
    \log \tilde P(y_k|\q,\e)= \gamma + \alpha_k \log P(y_k|\q,\e) + \beta_k
\end{equation}
where $P(y_k|\q,\e)$ is given by \eref{eq:post_from_llm},
 $\alpha$ and $\beta$ are parameters, and the value of $\gamma$ is determined so that $\sum_{k=1}^K \tilde P(y_k|\q,\e)=1$. That is, 
\begin{equation}\label{eq:gammai}
\gamma = -\log \sum_{k'=1}^K P(y_{k'}|\q,\e)^{\alpha_k} e^{\beta_{k'}}
\end{equation}

The $\alpha_k$ and $\beta_k$ parameters are estimated by minimizing the cross-entropy on a training set  $\mathcal{C}^\mathrm{train}=\{(\q^{(1)},y^{(1)}),\ldots,\allowbreak(\q^{(N)},y^{(N)})\}$ where $\q^{(i)}$ and $y^{(i)}$ are the query and the class of sample $i$:
\begin{equation}
    \mathcal{L} = -\frac{1}{N} \sum_{i=1}^{N} \log \tilde P(y^{(i)}|\q^{(i)},\e)
\end{equation}
In this work we take $\alpha_k$ to be a scalar, independent of the class. This is what is usually done for calibration~\citep{brummer2006,guo2017,platt1999}. In particular, temperature scaling~\citep{guo2017}, one of the most widely used calibration methods, corresponds to taking $\beta_k = 0$ for all $k$ and $\alpha_k$ a single scalar.

If we restrict the calibration transformation to have $\alpha_k=1$ and set the derivative of the cross-entropy to zero, we can derive the following expression for $\beta_k$:
\begin{equation} \label{eq:beta}
 \beta_k \! = \! \log \! \frac{N_k}{N} -  \log \! \left[ \frac{1}{N} \! \sum_{i=1}^N P(y_k|\q^{(i)},\e) e^{\gamma^{(i)}} \! \right]
 \end{equation}
 where $\gamma^{(i)}$ is given by \eref{eq:gammai} with $\q = \q^{(i)}$.
See Appendix for a derivation of this expression. Note that this is not a closed-form expression for $\beta_k$ but rather a system of equations since the right-hand side contains all the $\beta_k$ within the $\gamma^{(i)}$. 

We can now compare \eref{eq:new_post} and \eref{eq:cal_post}. Taking the logarithm of \eref{eq:new_post} for one specific class $k$ we get:
\begin{equation}\label{eq:log_new_post}
    \log \hat P(y_k|\q,\e)= \gamma' + \log P(y_k|\q,\e) + \beta'_k
\end{equation}
where $\gamma' = \log \delta$ and 
\begin{equation}\label{eq:beta_prime}
\beta'_k = \log \hat P(y_k|\q) - \log P(y_k|\q)
\end{equation}
The form of this expression is identical to that of \eref{eq:cal_post} when taking $\alpha_k=1$ for all $k$. Both $\gamma$ and $\gamma'$ are determined so that the posterior on the left-hand side adds to one. Hence, if $\beta_k = \beta'_k$, the two posteriors would be identical. 

Comparing the expressions for $\beta_k$ and $\beta'_k$ we can see that they coincide if we take $\hat P(y_k|\q) = N_k/N$ as we assume in our experiments for the SUCPA approach, and if we take $\gamma^{(i)} = 0$, since in that case the second term in $\beta_k$ coincides with \eref{eq:prior_approx}. Of course, $\gamma^{(i)}$ is not necessarily zero. Yet, as we will see in the experiments, making this assumption has little effect in the results. Nevertheless, we can also estimate the $\beta_k$ that satisfies \eref{eq:beta} exactly using an iterative approach where we first set $\gamma^{(i)} = 0$ and compute $\beta_k$ for all $k$, plug those values into \eref{eq:gammai} to get a new value for $\gamma^{(i)}$ and plug that back into \eref{eq:beta}, repeating these steps until convergence. We find that this algorithm leads to identical results as running linear logistic regression with $\alpha_k=1$. In the following, we will refer to the UCPA (and SUCPA) approach described in section \ref{sec:ucpa} as ``UPCA-naive'' (and ``SUCPA-naive''), whereas the iterative version of this method will be called simply UCPA (and SUCPA). \fref{fig:architecture} summarize both approaches for both the UCPA and SUCPA variants.

\section{Experimental Set Up}\label{sec:exp_setup}

\begin{table*}[t!]
\centering
\begin{tabular}{p{.1\textwidth}p{.18\textwidth}p{.07\textwidth}p{.54\textwidth}}
\hline
\textbf{Dataset} & \textbf{Class Priors} & \textbf{\parbox[t]{.07\textwidth}{Test \\ Samples}} & \textbf{Prompt Template}\\
\hline
TREC & 
\parbox[t]{.18\textwidth}{
0.28: Description \\
0.23: Number \\
0.19: Entity \\
0.16: Location \\
0.13: Person \\
0.02: Abbreviation
}
& 500 & 
``\textit{Classify the questions based on whether their answer type is a Number, Location, Person, Description, Entity, or Abbreviation.}

\textit{Question: }\texttt{[example 1]} \textit{Answer Type: }\texttt{[label 1]} \newline
$\ldots$ \newline
\textit{Question: }\texttt{[example n]} \textit{Answer Type: }\texttt{[label n]}

\textit{Question: }\texttt{[query]} \textit{Answer Type:''}
\\[.3em]\hline
SST-2 & 
\parbox[t]{.18\textwidth}{
0.50: Negative \\
0.50: Positive
}
& 1821 & 
``\textit{Review: }\texttt{[example 1]} \textit{Sentiment: }\texttt{[label 1]}\newline
$\ldots$ \newline
\textit{Review: }\texttt{[example n]} \textit{Sentiment: }\texttt{[label n]}

\textit{Review: }\texttt{[query]} \textit{Sentiment:''}
\\[.3em]\hline
AGNews & 
\parbox[t]{.18\textwidth}{
0.27: Technology \\
0.26: Business \\
0.25: Sports \\
0.22: World
}
& 1000 & 
``\textit{Classify the news articles into the categories of World, Sports, Business, and Technology.}

\textit{Article: }\texttt{[example 1]} \textit{Answer: }\texttt{[label 1]}\newline
$\ldots$ \newline
\textit{Article: }\texttt{[example n]} \textit{Answer: }\texttt{[label n]}

\textit{Article: }\texttt{[query]} \textit{Answer:''}
\\[.3em]\hline
DBpedia &
\parbox[t]{.2\textwidth}{
0.09: Artist \\
0.09: Nature \\
0.08: Athlete \\
0.08: Plant \\
0.08: Company \\
0.07: School \\
0.07: Village \\
0.07: Animal \\
0.07: Transportation \\
0.07: Politician \\
0.07: Album \\
0.06: Book \\
0.06: Building \\
0.05: Film \\

}
& 1000 & 
``\textit{Classify the documents based on whether they are about a Company, School, Artist, Athlete, Politician, Transportation, Building, Nature, Village, Animal, Plant, Album, Film, or Book.}

\textit{Article: }\texttt{[example 1]} \textit{Answer: }\texttt{[label 1]}\newline
$\ldots$ \newline
\textit{Article: }\texttt{[example n]} \textit{Answer: }\texttt{[label n]}

\textit{Article: }\texttt{[query]} \textit{Answer:''}
\\
\hline
\end{tabular}
\caption{Number of test samples, class priors in the test set, and prompt used for each dataset in this work. The instruction text for each case is taken from \cite{zhao2021calibrate}.}
\label{tab:datasets}
\end{table*}

We evaluate the proposed approach on the $n$-shot text classification task following a similar procedure as \citet{zhao2021calibrate}. We use four datasets for which the task is classification of a single text into known categories: binary sentiment analysis using SST-2~\citep{sst2}, 6-way question classification using TREC~\citep{trec}, 4-way news classification using AGNews~\citep{agnews}, and the 14-way ontology classification using DBPedia~\citep{dbpedia}. 
Table \ref{tab:datasets} shows the number of test samples, class priors and prompt used for each dataset.

We used the standard train and test partitions for all datasets. For AGNews and DBPedia we selected 1000 random samples from the test set since their test splits were too large for computation in our infrastructure. This approach was similar to the one used by \citet{zhao2021calibrate} where they selected 300 test samples (see their github repository\footnote{\url{https://github.com/tonyzhaozh/few-shot-learning}}). Also following  this work, we used GPT-2 XL  which has 1.5B parameters and consist of a decoder-only transformer architecture~\citep{transformer}. The checkpoint was downloaded from the \texttt{huggingface} website\footnote{\url{https://huggingface.co/gpt2-xl}}, and the code used to run experiments will be released upon paper acceptance.

\begin{figure*}[t!]
    \centering
   \includegraphics[width=\textwidth]{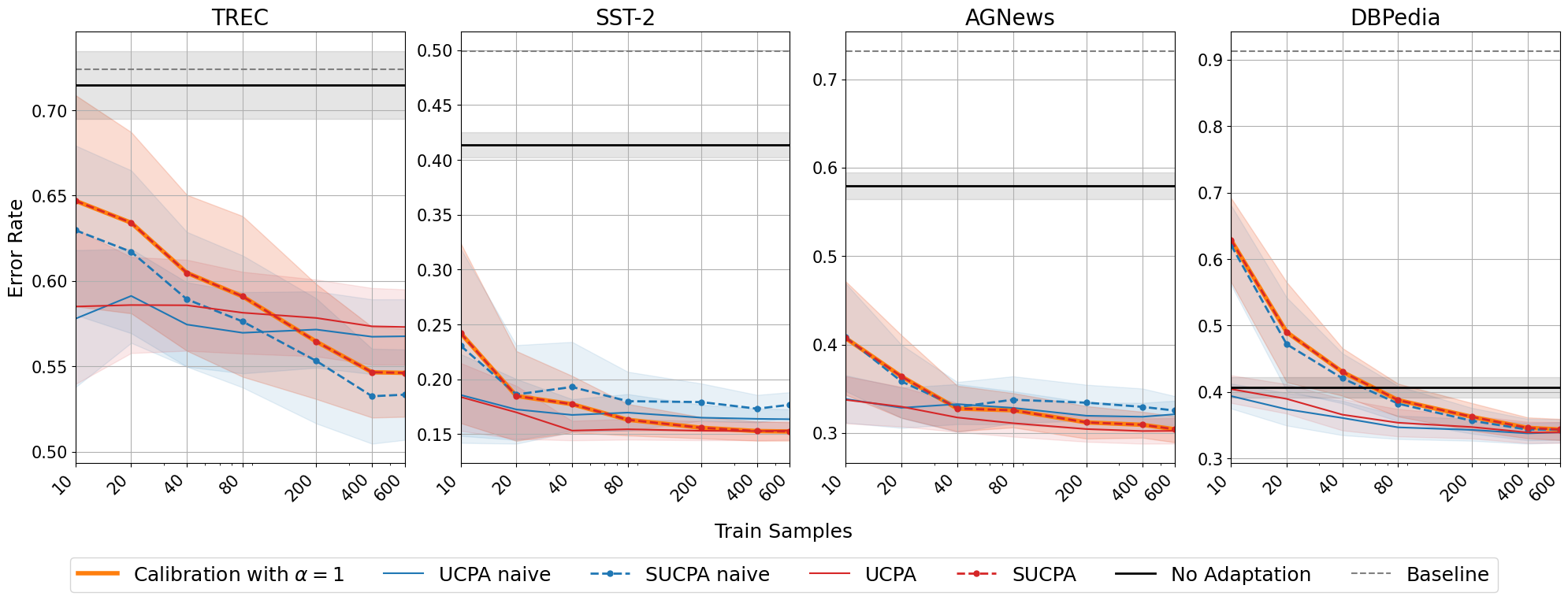}
    \caption{Error rate vs. the number of training samples used in the prior adaptation process in a zero-shot configuration. Red lines show the iterative approach for UCPA and SUCPA, whereas blue lines show the naive version. Orange curve shows the calibration results for $\alpha = 1$. Black curve shows the results without adaptation and grey dotted line represents the majority-class classifier (both are constant because they do not use training data).}
    \label{fig:num_samples}
\end{figure*}

\begin{figure*}[t!]
    \centering
   \includegraphics[width=\textwidth]{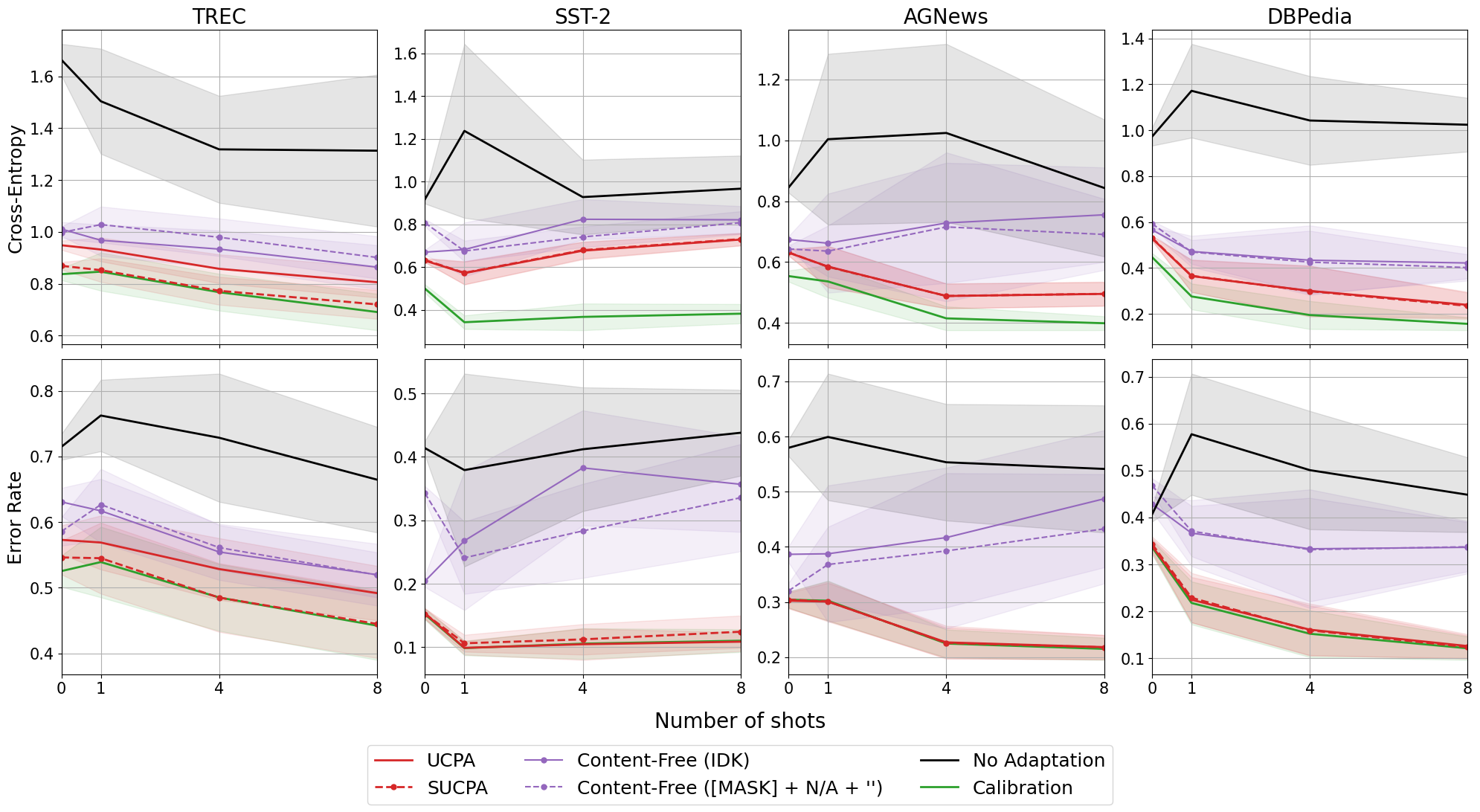}
    \caption{Cross-Entropy and Error Rate (1-Accuracy) vs. the number examples (shots) contained in the prompt for 600 training samples. Red lines show the iterative approach for UCPA and SUCPA. Lines in purple show the results for content-free adaptation and green line is the calibration using parameters $\alpha$ and $\beta$. As before, black line shows the case for which no adaptation has been performed.}
    \label{fig:num_shots}
\end{figure*}

For each dataset, a set of $n$ shots were selected by random sampling the train split and added to the preface (see Table \ref{tab:datasets}). Then, the $Q^{\mathrm{train}}$ set was generated by random sampling 600 samples from $Q^{\mathrm{train}}$ after discarding the samples added to the preface. The $\mathcal{C}^{\mathrm{train}}$ set to train the calibrator was the same as $Q^{\mathrm{train}}$ with the difference that $\mathcal{C}^{\mathrm{train}}$ contains the labels and $Q^{\mathrm{train}}$ does not. For some of the experiments we further subset the training set to smaller sizes. When doing this, we use 10 different seeds to generated the subsets to assess the variation in results due to varying training sets. 
For each training set we obtain posteriors on the test set and generate 100 bootstrap samples~\citep{bootstrapefron,bootstrapbengio}. The curves in the figures \ref{fig:num_samples} and \ref{fig:num_shots} show the mean performance on the pooled performance estimated from all training sets and test bootstraps and confidence intervals plotted one standard deviation away from the mean. 

We show results in terms error rate (1-accuracy), equivalently to \citet{zhao2021calibrate}. Further, in the final results, we also include the cross-entropy performance. Cross-entropy is a proper scoring rule which means that it evaluates the performance of the provided scores as posterior probabilities~\citep{spsr}. In the figures we show normalized cross-entropy, where the cross-entropy is divided by the cross-entropy of a naive system that always outputs the prior distribution, ignoring the input sample. A normalized cross-entropy larger than 1.0 indicates that the system is so badly calibrated that its performance is worse than that of a naive system~\citep{brummerthesis,ferrerarxiv}. 

\section{Results and Discussion}

We divide the results in two parts. First, we show the performance of the UCPA and SUCPA methods in comparison with supervised calibration and as a function of the number of training samples. Then, we study the performance of our proposed methods and the baseline methods as a function of the number of training shots in the prompt. 

\subsection{Effect of the training set size}\label{ssec:num_samples}

\fref{fig:num_samples} shows the error rate for all datasets as a function of the number of training samples used to perform domain adaptation for the 0-shot scenario (no examples added to the preface). This set is used either to train the calibration model (in which case the class labels are used), to compute \eref{eq:prior_approx} for UCPA/SUCPA-naive, and to compute \eref{eq:beta} for UCPA/SUCPA. For SUCPA, the training labels are used to compute $N_k/N$ which is used to obtain $\hat P(y|\e)$ and in \eref{eq:beta}. For UCPA, the training labels are never used and $N_k/N$ is assumed uniform over the classes. The figure also shows the results for the baseline system which takes the posteriors from \eref{eq:post_from_llm} without any prior adaptation. Finally, we show the performance of a naive baseline that always chooses the most likely class in the training data. 


We first note that, as explained in Section \ref{sec:calibration}, SUCPA and linear logistic calibration with $\alpha=1$ give identical performance. We found the same results for the case of 1, 4 and 8-shot learning, indicating that our iterative algorithm for estimating the $\beta_k$ that satisfies \eref{eq:beta} is working as expected.
We can also see for both UCPA and SUCPA (i.e., regardless of how the in-domain priors are estimated), the naive and the iterative approaches give similar performance, indicating that the average posterior in \eref{eq:prior_approx} is a  good approximation for the model's prior. Both proposed approaches show better performance than the original model for three of the four datasets even when very few (as low as 10) training samples are available to do the prior adaptation. In the case of DBPedia, however, we see that the SUCPA methods degrade performance compared to the baseline when the number of training samples is smaller than 80. This is due to the fact that the priors estimated as $N_k/N$ cannot be robustly estimated on so few samples for this 14-class.  Since the priors in this dataset are close to uniform (see Table \ref{tab:datasets}), in this case it is better to assume them uniform than to estimate them from a very small dataset.
A similar trend can be found for SST-2 and AGNews for which the priors are perfectly uniform so that assuming them is always better than estimating them from data. On the other hand, for the TREC dataset we can see that, given enough training samples, SUCPA works better than UCPA when the test priors are not uniform.
Similar trends as those seen in this figure were found for a prompt containing 1, 4 and 8 shots.


\subsection{Effect of the number of shots}\label{ssec:num_shots}

\fref{fig:num_shots} shows the effect of the number of examples (shots) added to the preface for each dataset in terms of error rate and cross-entropy when the number of training samples is 600. 
We compare our proposed methods with four systems: 1) the non-adapted posteriors (solid black line), 2)
the affine calibration method (solid green line) in which parameters $\alpha$ and $\beta$ are trained using the labelled training data, and 3) and 4) the two content-free calibration methods explained in Section \ref{sec:content_free} with the  two sets of content-free inputs that were used in the authors' code (see footnote above), namely, \{`IDK'\} and \{`[MASK]', `N/A', `'\'\}. 
We can see that, for 600 training samples, our proposed methods consistently improve upon the content-free baseline, as well as over the non-adapted posteriors. In most cases, the non-adapted model presented a mean cross-entropy close or higher than 1 for all number of shots, which implies that the model is useless for this task and cannot learn from the prompt shots. They also tend to have small standard deviation and a more stable tendency across the number of shots. Of course, the content-free approaches have the advantage that they do not require training data at all. Yet, these results show that, if unlabelled training data is available, we can obtain significant gains from our proposed UCPA approach. Further, if an estimate of the class priors is available, the SUCPA approach can lead to additional gains in datasets with imbalanced priors, like TREC.

\fref{fig:num_shots} also shows that the affine calibration system performs consistently better than our proposed approaches, particularly in terms of cross-entropy which better highlights issues of miscalibration compared to accuracy. This is expected since the calibrator is taking advantage of the labelled training data. Note that affine calibration is one specific case of a downstream classifier, one with very few parameters which can be trained with a small number of samples. A more complex downstream classifier may give further improvements, but would require larger labelled training datasets. 

With some exceptions (like in SST-2), increasing the number of examples added to the preface leads to gains in all methods that do some kind of adaptation. The original posteriors, on the other hand, have erratic behavior as a function of the number of shots with a very large standard deviation resulting from the specific selection of the examples.

Appendix shows the results when the number of training samples is set to 40. The trends are similar to those in \fref{fig:num_shots} with the exception that SUCPA works worse than UCPA in most cases due to a bad estimate of the class priors. In practice, the class priors would be estimated from knowledge of the task rather than from the training dataset, so that the SUCPA performance would in fact depend on how good that estimate is.

Overall, we can see that, in a scenario where a relatively small amount of unlabelled data is available for training, the proposed method leads to a large gain with respect to the non-adapted posteriors. 

\section{Conclusion and Future Work}

In this work we proposed a method for calibrating the posteriors generated by an LLM for a certain text classification task. We assume that only a relatively small number of unlabelled in-domain samples are available for adaptation. We propose a simple method for calibrating the posteriors generated by the LLM  by adapting the prior class distribution to the task of interest in an unsupervised manner. Optionally, the method allows the new priors to be set to the ones we expect to see during deployment, when known. When such priors are unknown, they can be assumed uniform. We show that, as long as the test priors can be estimated reasonably well, or that the uniform assumption is not too far off from the test distribution, the proposed approach works significantly better than the un-adapted posteriors even with a small amount of available adaptation samples. Further, we show that it works better than a previous approach where calibration is performed without using any adaptation data.


Given the results obtained in this work, additional experiments can be performed in the future. In the first place, a comparison could be done between this kind of domain adaptation and more complex techniques like (full or selective) finetuning. In addition, we would like to extend this method to more elaborated tasks that involves text generation like question answering and summarization. Further, a more detailed analysis should be done in terms of the model size, and newer LLMs should be tested in addition of GPT-2 XL. Finally, given that this is a general and principle-based method, we would like to investigate an extension of this work to other AI fields like computer vision and speech processing.



\bibliographystyle{acl_natbib}
\bibliography{references}

\ifbool{showappendix}{
    \clearpage






\appendix
\setcounter{page}{1}

\section{Affine Calibration with $\alpha=1$}\label{apx:beta_optim}

In this appendix we derive an expression for $\beta$ for logistic regression calibration when fixing $\alpha=1$ (see Section 4 in the main paper). That is, we assume the following expression for the calibrated posterior:
\begin{equation}\label{eq:cal_post_apx}
    \log \tilde P(y_k|\q,\e)= \gamma + \log P(y_k|\q,\e) + \beta_k
\end{equation}
where the $\gamma$ is simply a scaling factor so that the resulting posteriors add to 1 (see Section 4 for the expression for $\gamma$).

Given a set $\mathcal{C}^\mathrm{train}=\{(\q^{(1)},y^{(1)}),\ldots,\allowbreak(\q^{(N)},y^{(N)})\}$ where $\q^{(i)}$ and $y^{(i)}$ are the query and the class of sample $i$, the logistic regression approach estimates the $\beta_k$ parameters as the values that minimize
\begin{align}
    \mathcal{L} = -\frac{1}{N} \sum_{i=1}^{N} \log \tilde P(y^{(i)}|\q^{(i)},\e).
\end{align}
To obtain an expression for the $\beta_k$ we can set to zero the derivative $\frac{\partial \mathcal{L}}{\partial\beta_k}$:
\begin{align}
    \frac{\partial \mathcal{L}}{\partial\beta_k} &=\nonumber 
    \frac{\partial }{\partial\beta_k}\left( -\frac{1}{N} \sum_{i=1}^{N} \log \tilde P(y^{(i)}|\q^{(i)},\e)\right) \\[.5em] &=\nonumber
    - \frac{1}{N} \sum_{i=1}^N \frac{\partial }{\partial\beta_k} \log \tilde P(y^{(i)}|\q^{(i)},\e) \nonumber\\
    &=- \frac{1}{N} \sum_{i=1}^N \left( \frac{\partial }{\partial\beta_k} \gamma^{(i)} + \mathbbm{1}_{\{y^{(i)}=y_k\}}\right)\label{eq:partial_pre_gamma}
\end{align}
where $\mathbbm{1}_{\{\cdot\}}$ is the indicator function and $\gamma^{(i)}$ is defined in equation \ref{eq:gammai}.
Then,
\begin{align*}  
    \frac{\partial \mathcal{L}}{\partial\beta_k}\! &=\! -\frac{1}{N} \!\sum_{i=1}^N \!\left( \!
    -\! e^{\gamma^{(i)}} \! P(y_k|\q^{(i)}\!,\!\e)e^{\beta_k}\! +\!
    \mathbbm{1}_{\{y^{(i)}\!=\!y_k\}} \! \right) \\[.5em] &=
    -\frac{1}{N} \sum_{i=1}^N \left(
    - \tilde P(y_k|\q^{(i)},\e) +
    \mathbbm{1}_{\{y^{(i)}=y_k\}} \right). 
\end{align*}
Setting this derivative to zero, we get:
\begin{equation}
    \frac{1}{N} \sum_{i=1}^N \tilde P(y_k|\q^{(i)},\e) = \frac{N_k}{N}
\end{equation}
where $N_k$ is the number of samples that belongs to class $k$. Using Equation \eref{eq:cal_post_apx}, we obtain the expression for the optimal $\beta_k$
\begin{align*}
    & \frac{1}{N} \sum_{i=1}^N P(y_k|\q^{(i)},\e)e^{\beta_k}e^{\gamma^{(i)}} = \frac{N_k}{N} \\
    & e^{\beta_k} \frac{1}{N} \sum_{i=1}^N P(y_k|\q^{(i)},\e)e^{\gamma^{(i)}} = \frac{N_k}{N} \\
    & \beta_k  = \log \frac{N_k}{N} - \log \left[ \frac{1}{N} \sum_{i=1}^N P(y_k|\q^{(i)},\e)e^{\gamma^{(i)}}\right] 
\end{align*}

\section{Additional Results}\label{apx:results}
\fref{fig:num_shots_40} shows the results when the number of training samples is set to 40. Some of the curves like the one corresponding to the cross-entropy of the calibrated model in the DBPedia dataset that could not be fully plotted because cross-entropy could not be computed for a class with zero train probability. This may occur because some classes were never seen in the train set that was used in the calibration process.

The trends are similar to those shown in section \ref{ssec:num_shots} with the exception that SUCPA works worse than UCPA in most cases due to a bad estimate of the class priors. In addition, there are some cases such as TREC, in which adaptation improves the performance of the model but it still present a cross-entropy close to 1, which means that performance remains close to random.

\begin{figure*}[t!]
    \centering
    \includegraphics[width=\textwidth]{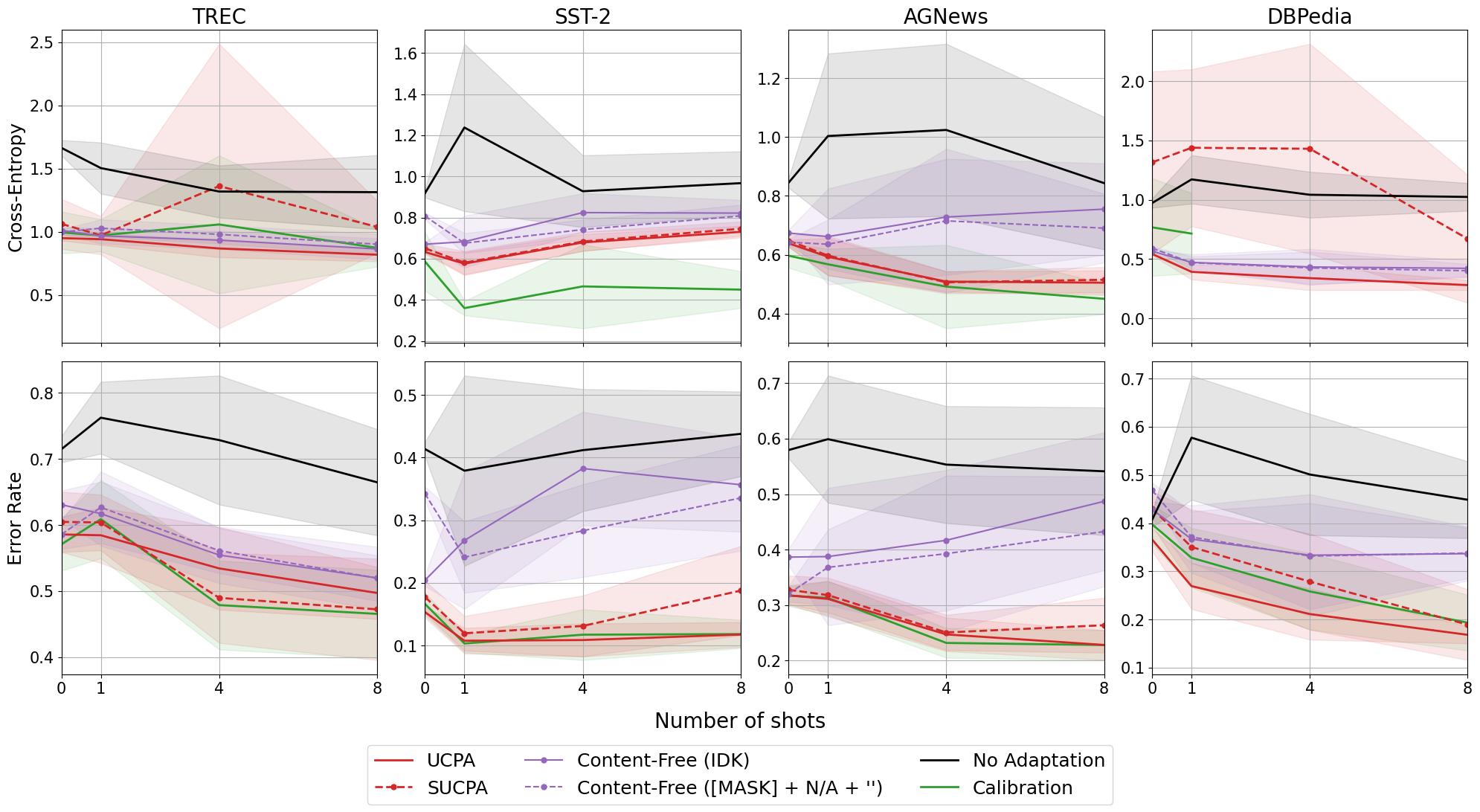}
    \caption{Cross-Entropy and Error Rate (1-Accuracy) vs. the number examples (shots) contained in the prompt for 40 training samples. Red lines show the iterative approach for UCPA and SUCPA. Lines in purple show the results for content-free adaptation and green line is the calibration using parameters $\alpha$ and $\beta$. As before, black line shows the case for which no adaptation has been performed.}
    \label{fig:num_shots_40}
\end{figure*}


}{}

\end{document}